\documentclass[letterpaper]{article}
\usepackage[]{aaai24} 
\usepackage{times} 
\usepackage{helvet} 
\usepackage{courier} 
\usepackage[hyphens]{url} 
\usepackage{graphicx} 
\usepackage{graphicx} 
\usepackage{natbib} 
\usepackage{caption} 
\usepackage{cite}
\usepackage{amsmath,amssymb,amsfonts}
\usepackage{graphicx}
\usepackage{textcomp}
\usepackage{xcolor}

\usepackage{soul}
\usepackage{amsthm}
\usepackage{algorithm}
\usepackage{algorithmicx}
\usepackage{algpseudocode}
\usepackage{caption}
\usepackage{subcaption}
\usepackage{multirow}
\usepackage{color}
\usepackage{enumitem}
\usepackage{booktabs}
\usepackage[figuresright]{rotating}
\usepackage{makecell}

\newcommand{\bL}{\ensuremath{\mathcal{L}}}

\newcommand{\bG}{\ensuremath{\mathcal{G}}}

\newcommand{\bC}{\ensuremath{\mathcal{C}}}

\renewcommand{\vec}[1]{\ensuremath{\mathbf{#1}}}
\newcommand{\stitle}[1]{\vspace{1mm} \noindent {\bf #1}}
\newcommand{\eg}{{\it e.g.}}

\newcommand{\ie}{{\it i.e.}}

\newcommand{\wrt}{w.r.t. }

\newcommand{\method}[1]{\textsc{#1}}
\newcommand{\model}{\method{HGPrompt}{}}
\newcommand{\pmodel}{\method{HGPrompt+}{}}

\newcommand{\eat}[1]{}
\def\corresAuthor{%
      \ifnum\value{eqfn}=1%
        \footnote{Corresponding authors.}%
        \setcounter{eqfn}{\value{footnote}}%
      \else%
        \footnotemark[\value{eqfn}]%
      \fi%
    }%

\def\equalcontribNew{%
      \ifnum\value{eqfn}=0%
        \footnote{Work was done while at Singapore Management University.}%
        \setcounter{eqfn}{\value{footnote}}%
      \else%
        \footnotemark[\value{eqfn}]%
      \fi%
    }%

\author{Xingtong Yu\textsuperscript{\rm 1}\equalcontribNew, Yuan Fang\textsuperscript{\rm 2}\corresAuthor, Zemin Liu\textsuperscript{\rm 3}, Xinming Zhang\textsuperscript{\rm 1}\corresAuthor}
\affiliations{
    \textsuperscript{\rm 1}University of Science and Technology of China, China\\
    \textsuperscript{\rm 2} Singapore Management University, Singapore\\
    \textsuperscript{\rm 3} National University of Singapore, Singapore\\
    yxt95@mail.ustc.edu.cn, yfang@smu.edu.sg, zeminliu@nus.edu.sg, xinming@ustc.edu.cn
}

\title{\model: Bridging Homogeneous and Heterogeneous Graphs \\ for Few-shot Prompt Learning
}

\begin{document}
\maketitle
\begin{abstract}
Graph neural networks (GNNs) and heterogeneous graph neural networks (HGNNs) are prominent techniques for homogeneous and heterogeneous graph representation learning, yet their performance in an end-to-end supervised framework greatly depends on the availability of task-specific supervision.
To reduce the labeling cost, pre-training on self-supervised pretext tasks has become a popular paradigm, but there is often a gap between the pre-trained model and downstream tasks, stemming from the divergence in their objectives.
To bridge the gap, prompt learning has risen as a promising direction especially in few-shot settings, without the need to fully fine-tune the pre-trained model.
While there has been some early exploration of prompt-based learning on graphs, they primarily deal with homogeneous graphs, ignoring the heterogeneous graphs that are prevalent in downstream applications. 
In this paper, we propose \model, a novel pre-training and prompting framework to unify not only pre-training and downstream tasks but also homogeneous and heterogeneous graphs via a \emph{dual-template} design. Moreover, we propose \emph{dual-prompt} in \model\ to assist a downstream task in locating the most relevant prior to bridge the  gaps caused by not only feature variations but also heterogeneity differences across tasks. 
Finally, we thoroughly evaluate and analyze \model\ through extensive experiments on three public datasets.
\end{abstract}

\section{Introduction}

Graph data is ubiquitous due to its ability to model relations between objects, such as bibliographic networks,
chemical compound graphs,
and social networks.
In particular, real-world graphs often involve multiple types of nodes and edges, 
such as authors, papers and conferences interacting in various ways in a bibliographic network. 
These graphs are called \emph{heterogeneous} graphs or heterogeneous information networks (HINs) \cite{sun2013mining, yang2020heterogeneous}, in contrast to conventional \emph{homogeneous} graphs that consist of only one type of node and edge.

On such homogeneous and heterogeneous graphs, graph neural networks (GNNs) \cite{kipf2016semi,wang2022searching,wang2023snowflake,li2023generalized} and heterogeneous graph neural networks (HGNNs) \cite{wang2019heterogeneous,lv2021we} have been widely deployed, respectively, for various tasks such as link prediction (LP), node classification (NC) and graph classification (GC). 
However, when trained in a supervised manner, GNNs and HGNNs
depend heavily on extensive task-specific labels, which are challenging or expensive to procure. Moreover, the supervised paradigm would require re-training for different tasks even on the same graph, incurring significant overheads.


Hence, the ``pre-train, fine-tune" paradigm \cite{dong2019unified} has emerged as a practical alternative to supervised learning. First, a pre-training step utilizes label-free graphs to learn task-irrelevant general properties on homogeneous \cite{hu2020strategies} or heterogeneous graphs  \cite{jiang2021contrastive}. Next, a fine-tuning step updates the pre-trained model to adapt it to downstream tasks using some task-specific labels. 
Apparently, the two steps optimize different objectives, creating a gap between pre-training and downstream tasks, which further leads to subpar performance \cite{liu2023pre}. In addition, fine-tuning a large pre-trained model can be costly, and still require considerable  task-specific labels to avoid overfitting. 

To address the issues with fine-tuning, prompt-based learning \cite{brown2020language,jia2021scaling} has become popular. Prompt learning seeks to extract the semantic relevance between a downstream task and the pre-trained model, acting as a bridge to align downstream objectives toward the pre-trained model. Under this approach, the parameters of the pre-trained model are frozen, and only a light-weight, task-specific prompt vector is tuned for a downstream task. Given much fewer learnable parameters in a prompt than in the pre-trained model, prompt-tuning can be both efficient and effective especially in few-shot tasks with very scarce task-specific labels.

Fueled by the success of prompt learning in language models, researchers have attempted its application to graphs \cite{liu2023graphprompt,DBLP:conf/kdd/SunCLLG23, sun2022gppt,DBLP:conf/kdd/TanGDL23, yu2023multigprompt}. Typically, these approaches first introduce a unification template to bridge pre-training and downstream tasks on graphs. Then, they design a learnable prompt that appends to or modifies the input 
of the downstream task, to better align the task with the pre-trained graph model. Yet these previous works are only designed for homogeneous graphs, and do not consider the gap brought by tasks on heterogeneous graphs.

\begin{figure}[t]
\centering
\includegraphics[width=0.99\linewidth]{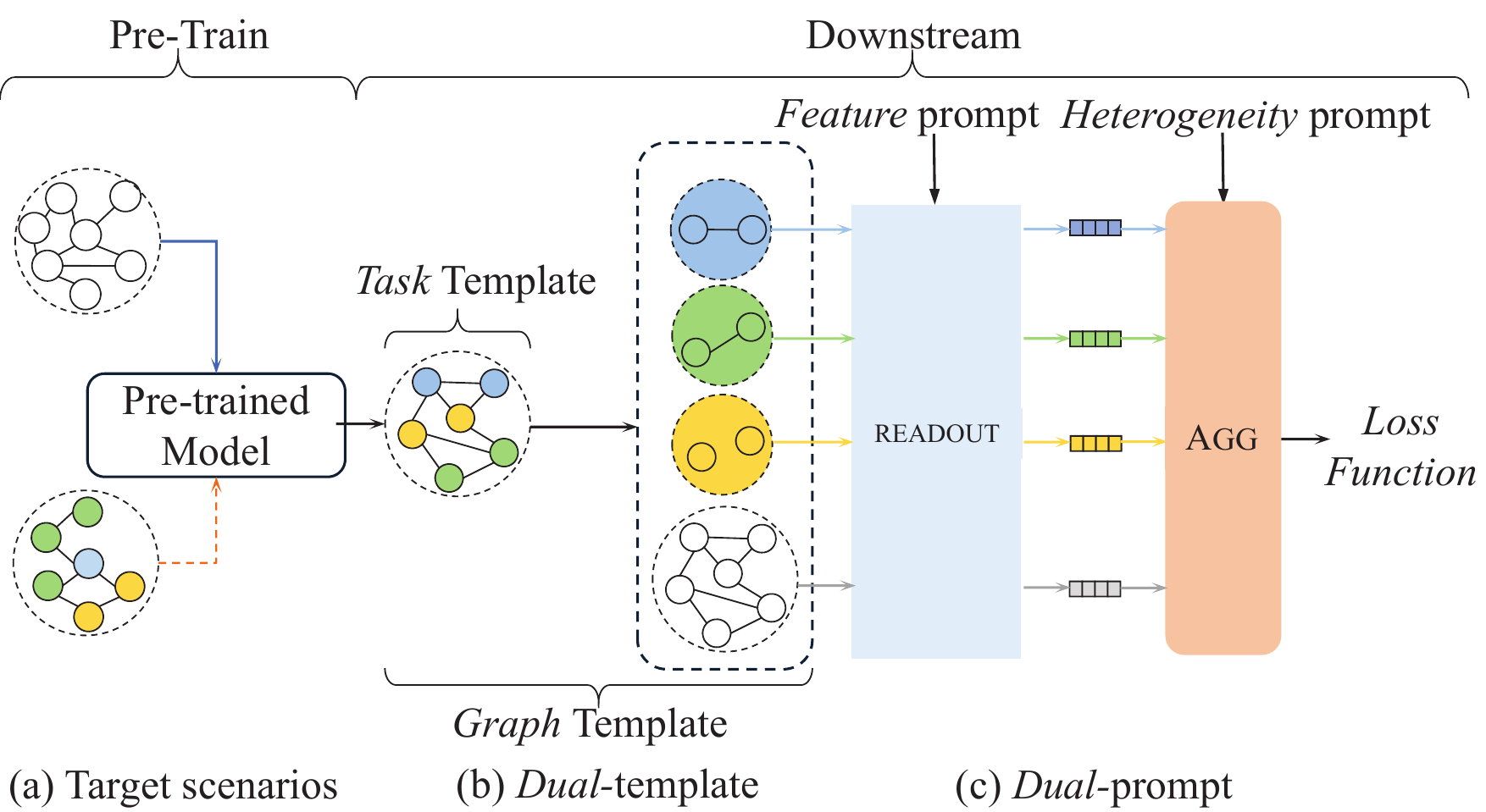}
\caption{Illustration of \model. Black-and-white graphs are homogeneous; colored graphs are heterogeneous, where colors indicate different types of nodes. }
\label{fig.intro-motivation}
\end{figure}

In this study, 
we present a novel few-shot prompt learning framework amenable to heterogeneous graphs, called \model. It aims to bridge the gap between homogeneous and heterogeneous graphs. 
Specifically, \model\ considers two target scenarios as shown in Fig.~\ref{fig.intro-motivation}(a): given a heterogeneous graph in a downstream task, we can leverage a pre-trained model that may be trained on either heterogeneous or homogeneous graphs. 
Note that the inconsistency between pre-training and downstream graphs could stem from several reasons.
For instance, the heterogeneity in pre-training graphs may be masked for privacy, or the pre-training graphs are not made available and we only have access to a pre-trained model on homogeneous graphs. In general, we may or may not have control over pre-training, and thus it is more flexible with the two target scenarios. However, the problem is non-trivial due to two key challenges. 


First, how do we \emph{unify downstream tasks on heterogeneous graphs with pre-training tasks irrespective of their graph heterogeneity?} In the language domain, the tasks are unified by a common template of masked language modeling \cite{brown2020language}. Similar attempts exist on graph-based tasks. For example, GraphPrompt \cite{liu2023graphprompt} converts popular graph-based tasks to a common template of predicting subgraph similarity. However, this does not deal with the heterogeneity differences in the input graphs to pre-training and downstream stages. Essentially, we aim to unify  not only the template of the tasks, but also the template of graphs, and propose a \emph{dual-template} design. On one hand, we simply follow GraphPrompt to unify link prediction, node classification and graph classification using a common task template based on subgraph similarity. On the other hand, in \model\ we propose a unified graph template to convert a heterogeneous graph into multiple homogeneous subgraphs, as illustrated in Fig.~\ref{fig.intro-motivation}(b). To be more specific, we sample one homogeneous subgraph for each type of node/edge, plus an additional one without considering their types. Hence, a heterogeneous graph is unified with homogeneous graphs, yet the heterogeneity is still discernible across graph samples. 


Second, with unified templates of both tasks and graphs, how do we \emph{design a prompt to narrow the gaps across tasks, caused by not only feature variations but also heterogeneity differences?} 
In the language domain, a prompt is simply a sequence of word tokens that reformulate the downstream task to narrow the gap with pre-training. Similarly, recent graph prompting methods \cite{liu2023graphprompt,sun2022gppt,DBLP:conf/kdd/TanGDL23} modify the input embeddings to close the task gap, whilst leaving the heterogeneity gap as an open question to solve. In \model,
we introduce a \emph{dual-prompt} as illustrated in Fig.~\ref{fig.intro-motivation}(c). First, a  feature  prompt modifies the input to the subgraph readout similar to GraphPrompt, based on the observation that different tasks may focus on different features \cite{liu2023graphprompt}.
Second, a  heterogeneity prompt further modifies the aggregation weights of the multiple homogeneous subgraphs that are converted from the input heterogeneous graph, as different tasks may focus on different facets of heterogeneity. 



To summarize, the contribution of this work is threefold. (1) We propose \model, a prompt learning framework on graphs, which bridges the gap between homogeneous and heterogeneous graphs across pre-training and downstream tasks. (2) In \model, we design a unified graph template for both heterogeneous and homogeneous graphs, and propose a dual-layer prompt to narrow the gaps caused by the differences in tasks and heterogeneity.  (3) We conduct extensive experiments on three benchmark datasets, demonstrating the advantages of \model.

\section{Related Work}

\stitle{Graph pre-training.}
Borrowing insights from pre-training methodologies in both vision \cite{bao2021beit} and language domains \cite{dong2019unified}, many GNN-based pre-training techniques have been introduced for graph data \cite{kipf2016variational,qiu2020gcc,bo2023graph}. Typically, these methods harness the inherent graph structures in a self-supervised fashion, facilitating the transfer of knowledge to downstream tasks through a fine-tuning step. There has also been a surge in the application of pre-training techniques to heterogeneous graphs \cite{jiang2021contrastive,jiang2021pre,yang2022self,hwang2020self}. Unlike their homogeneous counterparts, heterogeneous graphs are characterized by the presence of heterogeneous nodes and edges. Consequently, these techniques are tailored to leverage the intricate semantic relationships between nodes, originating from the inherent heterogeneity.
However, the inconsistency between pre-training and fine-tuning objectives can hinder the performance of downstream tasks \cite{liu2023pre}. While pre-training aims to extract inherent knowledge from the graph in a self-supervised manner, fine-tuning focuses on specific supervision tailored to downstream tasks. 

\stitle{Graph prompt learning.}
First emerged in the language domain \cite{brown2020language}, prompt learning seeks to reconcile the differences between pre-training and fine-tuning objectives. This is usually achieved by crafting task-specific prompts that guide downstream tasks, while freezing the pre-trained model weights. 
The success of this technique in language and vision \cite{jia2021scaling} has sparked interest in its adoption in graph learning. 
GPPT \cite{sun2022gppt} employs link prediction as its pre-training task, but its prompts are only designed for the downstream node classification task.
Both GraphPrompt \cite{liu2023graphprompt,yu2023generalized} and ProG \cite{DBLP:conf/kdd/SunCLLG23} aim to unify pre-training with multiple types of downstream tasks. On one hand, GraphPrompt employs subgraph similarity as its template and devises a learnable prompt for downstream few-shot learning. On the other hand, ProG transforms node and edge-level tasks into graph-level tasks, and assumes a  meta-learning setting for prompt learning.  Another work VNT \cite{DBLP:conf/kdd/TanGDL23} introduces a learnable virtual node prompt to assimilate task-specific information, while only focusing on downstream node classification in a meta-learning setup. 
However, none of these methods thoroughly address prompt learning toward bridging the gap between homogeneous and heterogeneous graphs. 

\section{Preliminaries and Problem Statement}

In this section, we present some preliminaries on heterogeneous graph and introduce our problem. 

\stitle{Graph preliminaries.}
A \emph{heterogeneous graphs}, also known as a heterogeneous information network \cite{sun2012mining}, is represented as $G=(V,E,A,R,\phi,\varphi)$, where $V$ is the set of nodes and $E$ is the set of edges, with a node type mapping function $\phi : V \rightarrow A$ and an edge type mapping function $\varphi : E \rightarrow R$ for a set of node and edge types denoted by $A$ and $R$, respectively, such that $|A| + |R| > 2$. We also assume an input feature matrix for the nodes, $\vec{X}\in\mathbb{R}^{|V|\times d}$. 

 A \emph{homogeneous graph} follows the same definition as the heterogeneous graph, except that $|A|=|R|=1$, \ie, there is only a single node and edge type.

\stitle{Problem definition.}
In our target scenarios, either homogeneous or heterogeneous graphs can be used for pre-training via a link prediction task. The use of link prediction in pre-training stems from the general availability of links in large-scale graphs without additional annotation costs \cite{hamilton2017inductive,liu2023graphprompt}.

For downstream tasks, we deal with heterogeneous graphs, and focus on two widely deployed tasks following GraphPrompt \cite{liu2023graphprompt}, namely, node classification and graph classification.
For node classification on a heterogeneous graph $G$, let $C$ be the set of node classes, such that a node in $G$ is associated with a class in $C$; 
for graph classification on a set of heterogeneous graphs $\bG$, let $\bC$ be the set of graph labels, such that a graph in $\bG$ is associated with a class in $\bC$. 
In both tasks, we aim to predict the unknown class labels for the instances in the test set of each task.
%
%
In particular, we consider a \emph{few-shot} setting: For each node or graph class, we only have $k$ labeled instances for learning.

\begin{figure*}[t]
\centering
\includegraphics[width=1\linewidth]{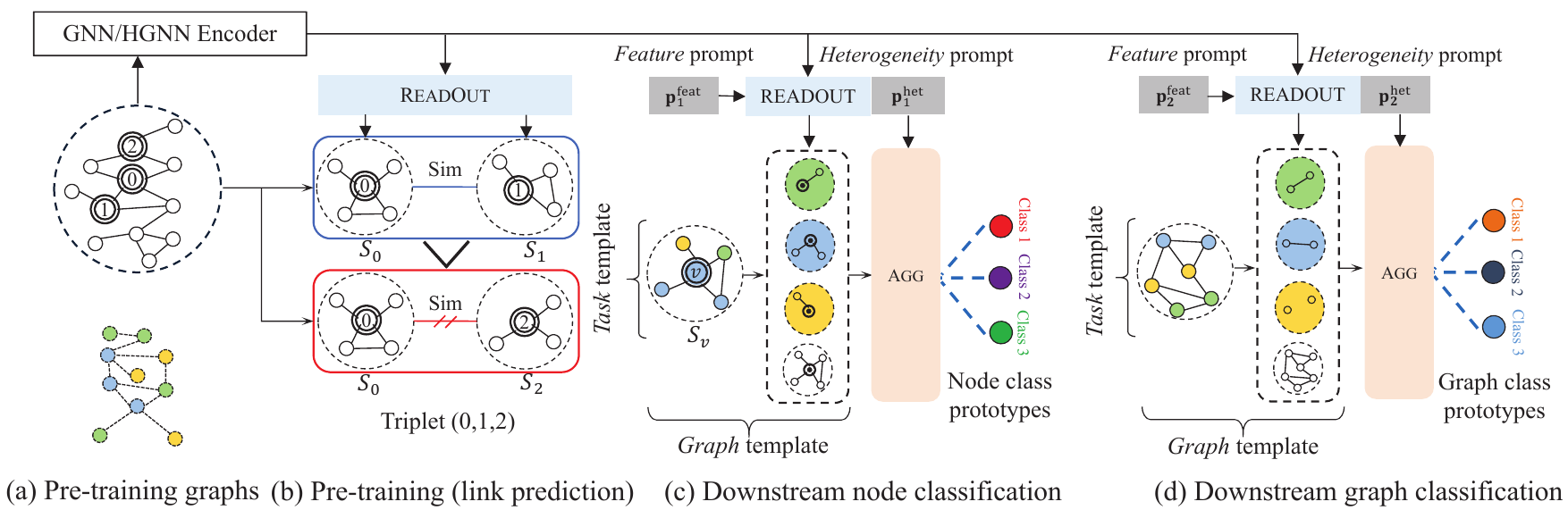}
\vspace{-4mm}
\caption{Overall framework of \model. (a) Pre-training graphs can be either homogeneous or heterogeneous. 
(b) Pre-training task with link prediction on a homogeneous graph$^\text{\ref{footnote:pre-train}}$. (c) Downstream node classification and (d) graph classification on heterogeneous graphs. 
Black-and-white graphs are homogeneous; colored graphs are heterogeneous, where colors indicate different types of nodes. }
\label{fig.framework}
\end{figure*}

\section{Proposed Model: \model}
In this section, we present our proposed model \model.

\subsection{Overall Framework}\label{sec:model:unification}

We begin with the overall framework of \model, as seen in Fig.~\ref{fig.framework}.
Given some homogeneous or heterogeneous graphs in Fig.~\ref{fig.framework}(a), we aim to pre-train a GNN model using link prediction in Fig.~\ref{fig.framework}(b), since links are available in all graphs ``for free'' without extra annotation costs. Here we assume the more general scenario of pre-training on homogeneous graphs\footnote{\label{footnote:pre-train}Pre-training on heterogeneous graphs can be achieved by also applying the graph template. See illustration in Appendix A.}. To unify pre-training and downstream stages, we propose a \emph{dual-template} design: first, a \emph{graph} template is applied to convert each heterogeneous graph to multiple homogeneous ones in Fig.~\ref{fig.framework}(c, d); second, a \emph{task} template is applied to convert various tasks to a common subgraph similarity prediction task. In the downstream stage, we further propose a \emph{dual-prompt} formulation consisting of both a \emph{feature} prompt and a \emph{heterogeneity} prompt to bridge the feature variations and heterogeneity differences across tasks.    

We will discuss the dual-template design as the basis of our unification (Sect.~\ref{sec:model:template}). Next, we present the pre-training stage (Sect.~\ref{sec:model:pre-train}) and introduce the dual-prompt design for the downstream stage (Sect.~\ref{sec:model:downstream}).

\subsection{Dual-Template Unification}\label{sec:model:template}

We introduce the basis of unifying pre-training and downstream tasks, namely, the graph and task templates.

\stitle{Graph template.}
We propose to unify the heterogeneity of the input graphs through graph template.
Our strategy is to convert a heterogeneous graph $G=(V,E,A,R,\phi,\varphi)$ into multiple homogeneous ones based on the node types\footnote{Edge types can also be considered just like node types. For brevity, our discussion will illustrate node types only.}. Specifically, given a node type $i \in A$, we can extract a homogeneous sugbraph $G^i=(V^i, E^i)$ from $G$ such that
\begin{align}
V^i &= \{v\in V \mid \phi(v)=i\}, \\
E^i &= \{(a,b) \in E \mid a \in V^i \land b\in V^i \}.
\end{align}
In this way, the heterogeneous $G$ can be converted into a set of homogeneous graphs $\{G^i:i\in A\}$, each retaining one aspect of the heterogeneity. To also preserve the interactions between different types, we further consider a  homogeneous graph $G^0=(V,E)$, \ie, it retains the full topology of $G$, but it does not distinguish the types. We call the process of converting a heterogenous graph into $|A|+1$ homogeneous graph as applying a \emph{graph template}, denoted by $\mathcal{GT}$:
\begin{align}
    \mathcal{GT}(G) = \{G^0\} \cup \{G^i:i\in A\}.
\end{align}

To summarize, the graph template unifies downstream heterogeneous graphs into the same format as homogeneous graphs that may be used in pre-training. When there are heterogeneous graphs in pre-training, the same graph template can be applied to them as well.


\stitle{Task template.} Next, to unify different tasks, we mainly follow GraphPrompt \cite{liu2023graphprompt} by converting different task instances into subgraphs, and leverage a common task template of predicting subgraph similarity. However, here we also need to account for the graph template in the formulation of subgraphs and their similarity computation. 

First, for node-level tasks including link prediction (LP) and node classification (NC), we can instantiate each node $v$ involved in LP or NC into a context subgraph $S_v$. A common strategy for $S_v$ is to employ a breadth-first search to extract a $\delta$-hop subgraph centering on $v$ \cite{liu2023graphprompt}. 
For graph classification (GC), as its instances are already in graph forms, \ie, $S_G=G$ for a graph instance $G$. 

Next, let $\mathbf{s}_v$ denote the embedding vector of a subgraph $S_v$. If $S_v$ is homogeneous, we can use a direct readout. If $S_v$ is heterogeneous, the graph template is applied to obtain a set of homogeneous graphs $\mathcal{GT}(S_v)$, which are individually readout and then aggregated into  $\mathbf{s}_v$. That is,
\begin{align}
\mathbf{s}_v = \textsc{Agg}(\{\textsc{ReadOut}(S_v^i)\mid S_v^i \in \mathcal{GT}(S_v)\}), \label{eq.agg.readout}
\end{align}
where the readout layer for a subgraph $S$ is further defined as an aggregation over the features of nodes in $S$:
\begin{align}
\textsc{ReadOut}(S)=\textsc{Agg}_2(\{\vec{h}_v:v\in V(S)\}),
\label{eq.readoutlayer}
\end{align}
where $\vec{h}_v$ denotes the embedding of node $v$ before the readout layer, and $V(S)$ is the set of nodes in $S$. 
Then, the three tasks can be reformulated into predicting the similarity between (sub)graph instances \cite{liu2023graphprompt}.
\begin{itemize}[leftmargin=*]
\item \textbf{\textit{Link prediction (LP)}}: Given a graph $G$ and a node triplet $(v,a,b)$ such that $(v,a)$ is an edge in $G$ and $(v,b)$ is not. It is expected that
\begin{align}
\text{sim}(\vec{s}_v,\vec{s}_a)>\text{sim}(\vec{s}_v,\vec{s}_b),
\end{align}
where $\text{sim}(\cdot,\cdot)$ is a similarity function such as cosine similarity in our implementation.

\item \textbf{\textit{Node Classification (NC)}}: 
Given a graph $G$ with a set of node classes $C$, and a labeled node set $D=\{(v_1,\ell_1),(v_2,\ell_2),\ldots\}$ such that $\ell_i$ is the class label of $v_i$.
In the context of $k$-shot learning, precisely $k$ pairs of $(v_i,\ell_i=c) \in D$ exist for every class $c \in C$.
For a class $c$, we construct a \emph{node class prototype}, which can be understood as a ``virtual'' subgraph. Its embedding can be computed as $\vec{\tilde{s}}_c = \frac{1}{k}\sum_{(v_i,\ell_i)\in D, \ell_i=c} \vec{s}_{v_i}$. 
Subsequently, for an unlabeled node $v_j$, its class label $\ell_j$ can be predicted as
\begin{align}
\ell_j = \textstyle\arg\max_{c\in C} \text{sim}(\vec{s}_{v_j}, \vec{\tilde{s}}_c).
\end{align}
\item \textbf{\textit{Graph classification (GC)}}: Given a collection of graphs $\mathcal{G}$ and a set of graph classes $\mathcal{C}$, accompanied by a set of labeled graphs $\mathcal{D}=\{(G_1,L_1),(G_2,L_2),\ldots\}$ such that $L_i$ is the class label of $G_i$. In $k$-shot learning, there exist precisely $k$ pairs of $(G_i,L_i=c) \in \mathcal{D}$ for every class $c\in \mathcal{C}$. For a class $c$, we also define a \emph{graph class prototype} which can be represented by 
$\vec{\tilde{s}}_c = \frac{1}{k}\sum_{(G_i,L_i)\in \mathcal{D}, L_i=c} \vec{s}_{G_i}.$
Then, given an unlabeled graph $G_j$, its class label $L_j$ can be predicted as 
\begin{align}
    L_j = \textstyle\arg\max_{c\in \mathcal{C}} \text{sim}(\vec{s}_{G_j}, \vec{\tilde{s}}_c).
\end{align}
\end{itemize}

\subsection{Pre-Training Task}\label{sec:model:pre-train}
We employ the LP task for pre-training, capitalizing on the abundance of links in large-scale graph data as self-supervision \cite{hamilton2017inductive,liu2023graphprompt}. 
As shown in Fig.~\ref{fig.framework}(b), consider a triplet (0, 1, 2) where (0, 1) is an edge and (0, 2) is not. It can be used to guide the computation of subgraph similarity toward $\text{sim}(\vec{s}_0,\vec{s}_1)>\text{sim}(\vec{s}_0,\vec{s}_2)$, based on the task template. In general, we gather many such triplets from the pre-training graphs to serve as self-supervision $\mathcal{D}_\text{pre}$. Then, the pre-training loss is defined as $\bL_{\text{pre}}(\Theta)=$
\begin{align} \label{eq.pre-train-loss}
    -\sum_{(v,a,b)\in\mathcal{D}_\text{pre}}\ln\frac{\exp\left(\frac{1}{\tau}\text{sim}(\vec{s}_v,\vec{s}_a)\right)}{\sum_{u\in\{a,b\}}\exp\left(\frac{1}{\tau}\text{sim}(\vec{s}_v,\vec{s}_u)\right)},
\end{align}
where $\tau$ is a temperature hyperparameter and $\Theta$ represents the model weights of GNN/HGNN. 

The pre-training stage outputs the trained model weights, \ie, $\Theta_0=\arg\min_\Theta \mathcal{L}_{\text{pre}}(\Theta)$,
which will be deployed in downstream tasks as we shall see next.

\subsection{Dual-prompting for downstream tasks}\label{sec:model:downstream}

We aim to address NC and GC tasks in downstream, while leveraging an LP-based pre-trained model.
Since the pre-training and downstream tasks involve different objectives, we turn to the paradigm of prompt learning to bridge the gap. In our scenario, different tasks may focus on different features, or different facets of heterogeneity in the input graph.  
Hence, we propose a \emph{dual-prompt} as shown in Fig.~\ref{fig.framework}(c,d), consisting of a \emph{feature} prompt to handle the feature variations and a \emph{heterogeneity} prompt to cope with the heterogeneity differences across different tasks. 

\stitle{Feature prompt.} In the language domain, a prompt modifies or reformulates the input to the pre-trained model based on the specific downstream task, to align the task better with the pre-trained model.
Similarly, on graphs, a straightforward way is to modify node features in the input or hidden layers of the pre-trained model. Here, we follow GraphPrompt \cite{liu2023graphprompt} to modify the node features before the readout layer in  Eq.~\eqref{eq.readoutlayer}, which enables different downstream tasks to focus on different sets of features. Specifically, on a given task, let $\vec{p}^\text{feat}$ be a task-specific learnable vector to serve as the feature prompt for the task. Under the prompt, the readout layer for some subgraph $S$ becomes  
\begin{align}\label{eq:prompt-fw}
\textsc{ReadOut}(\{\vec{p}^\text{feat}\odot\vec{h}_v\mid v\in V(S)\}),
\end{align}
where $\odot$ denotes the element-wise multiplication. In other words, the feature prompt modifies the feature importance based on the specific needs of the task. Note that the prompt $\vec{p}^\text{feat}$ has the same dimension as the node embeddings.



\stitle{Heterogeneity prompt.} 
When dealing with a heterogeneous (sub)graph $S$ downstream, we need to apply the graph template, which involves an extra layer to aggregate the individual readout outputs from each homogeneous subgraph in $\mathcal{GT}(S)$; see Eq.~\eqref{eq.agg.readout}. 
As various facets of heterogeneity (as suggested by different node/edge types) exist, to distinguish the importance of facets across different tasks, we propose a heterogeneity prompt to modify the input to the aggregation layer. Specifically, let $\vec{p}^\text{het}=(p^\text{het}_0,p^\text{het}_1,\ldots,p^\text{het}_{|A|})$ denote a task-specific learnable vector as the heterogeneity prompt for a particular task. Note that the prompt vector has $|A|+1$ dimensions, where each dimension $p^\text{het}_i$ intends to weigh the $i$-th subgraph from the graph template $\mathcal{GT}(\cdot)$. Under the prompt, for some subgraph $S$, the aggregation layer after applying the graph template becomes
\begin{align}\label{eq.prompt-hetw}
  \hspace{-1mm}  \textsc{Agg}(\{(1+p^\text{het}_i) \odot \textsc{ReadOut}(S^i)\mid S^i \in \mathcal{GT}(S)\}),
\end{align}
which gives the embedding vector $\vec{s}$ for the subgraph $S$.
Hence, the prompt is able to adjust the importance of every facet carried by each subgraph $S^i \in \mathcal{GT}(S)$ based on the task nature. 

\stitle{Prompt tuning.}
To tune the dual-prompt for a downstream task, we resort to a loss based on subgraph similarity just as pre-training. 
Consider an NC or GC task with a labeled training set $\mathcal{D}_\text{down}=\{(x_1,y_1),(x_2,y_2),\ldots\}$,
where each $x_i$ is either a node or a graph, and $y_i\in Y$ is $x_i$'s class label from a set of classes $Y$. Then, the prompt tuning loss is defined as $\bL_{\text{down}}(\vec{p}^\text{feat},\vec{p}^\text{het})=$
\begin{align}
    -\sum_{(x_i,y_i)\in \mathcal{D}_\text{down}}\ln\frac{\exp\left(\frac{1}{\tau}\text{sim}(\vec{s}_{x_i},\tilde{\vec{s}}_{y_i})\right)}{\sum_{c\in Y}\exp\left(\frac{1}{\tau}\text{sim}(\vec{s}_{x_i},\tilde{\vec{s}}_{c})\right)}.
\end{align}
Note that the subgraph and class prototype embeddings $\vec{s}_{x_i}$ and $\tilde{\vec{s}}_c$ are generated based on the prompt vectors $\vec{p}^\text{feat},\vec{p}^\text{het}$; see Eqs.~\eqref{eq:prompt-fw} and \eqref{eq.prompt-hetw}. During prompt tuning, only the light-weight prompt vectors are tuned, while the pre-trained weights $\Theta_0$ are frozen without any fine-tuning. Such parameter-efficient tuning is amenable to few-shot settings when $\mathcal{D}_\text{down}$ only consists of a few training examples. 




\section{Experiments}

In this section, we conduct experiments to evaluate our proposed approach, and analyze the empirical results.
\begin{table}[tbp]
\center
\small
\caption{Summary of datasets.}
\vspace{-1mm}
 \label{table.datasets}
\resizebox{1\linewidth}{!}{%
\begin{tabular}{@{}c|rcrccc@{}}
\toprule
	& \makecell[c]{\# Nodes} & \# \makecell[c]{Node \\ Types} & \# \makecell[c]{Edges} & \# \makecell[c]{Edge \\ Types} & \makecell[c]{Target \\ Type} & \# \makecell[c]{Classes} \\
\midrule
     ACM & 10,942 & 4 &  547,872 & 8 & paper & 3\\ 
     DBLP &  26,128 & 4 & 239,566 & 6 & author & 4\\
     Freebase &  180,098 & 8 &  1,057,688 & 36 & book & 7\\    
 \bottomrule
\end{tabular}}
\end{table}

\begin{table*}[tbp] 
    \centering
    \small
     \addtolength{\tabcolsep}{-1mm}
    \caption{Evaluation of node and graph classification. The best method is bolded and the runner-up is underlined.
    }
    \vspace{-1mm} 
    \label{table.node-graph-classification}%
    \resizebox{1\linewidth}{!}{%
    \begin{tabular}{@{}l|cc|cc|cc|cc|cc|cc@{}}
    \toprule
   \multirow{3}*{Methods} &\multicolumn{6}{c|}{Node classification} &\multicolumn{6}{c}{Graph classification}\\
   & \multicolumn{2}{c}{ACM} & \multicolumn{2}{c}{DBLP} & \multicolumn{2}{c|}{Freebase}
   & \multicolumn{2}{c}{ACM} & \multicolumn{2}{c}{DBLP} & \multicolumn{2}{c}{Freebase}\\ 
      &\multicolumn{1}{c}{MicroF (\%)} &\multicolumn{1}{c}{MacroF (\%)}
      &\multicolumn{1}{c}{MicroF (\%)} &\multicolumn{1}{c}{MacroF (\%)}
      & MicroF (\%) & MacroF (\%)
      &\multicolumn{1}{c}{MicroF (\%)} &\multicolumn{1}{c}{MacroF (\%)}
      &\multicolumn{1}{c}{MicroF (\%)} &\multicolumn{1}{c}{MacroF (\%)}
      & MicroF (\%) & MacroF (\%)

      \\\midrule\midrule
    \method{GCN} 
    & 50.26$\pm$12.17 & 44.21$\pm$16.62 
    & 51.36$\pm$15.40 & 48.62$\pm$17.27 
    & 17.11$\pm$11.80 & 15.35$\pm$9.54
        & 33.56$\pm$\ \ 1.05 & 17.72$\pm$\ \ 2.78 
    & 38.12$\pm$15.02 & 35.01$\pm$16.24 
    & 17.38$\pm$2.94 & 16.50$\pm$2.54
\\ 
    \method{GAT} 
    & 38.50$\pm$\ \ 7.86 & 28.01$\pm$12.44 
    & 65.04$\pm$11.68 & 62.83$\pm$13.21 
    & 17.93$\pm$\ \ 8.50 & 16.51$\pm$6.81
        & 33.52$\pm$\ \ 1.05 & 17.25$\pm$\ \ 1.53 
            & 46.47$\pm$14.51 & 40.33$\pm$16.84 
    & 16.30$\pm$2.69 & 16.01$\pm$2.14
\\\midrule
    \method{Simple-HGN}
    & 45.57$\pm$10.64	& 40.58$\pm$14.21	
    & 59.74$\pm$15.22	& 57.34$\pm$16.29	
    & 17.85$\pm$\ \ 9.12	& 16.00$\pm$7.88
        & 33.30$\pm$10.87	& 16.83$\pm$\ \ 6.88	
            & 44.56$\pm$15.80	& 39.52$\pm$17.12
    & 17.83$\pm$2.81	& 16.71$\pm$2.49
\\
    \method{HAN}
    & 62.22$\pm$\ \ 9.11	& 57.19$\pm$11.79	
    & 62.46$\pm$18.92	& 60.64$\pm$19.73	
    & 18.73$\pm$12.40	& 16.81$\pm$8.16
        & 35.35$\pm$\ \ 8.93	& 22.27$\pm$\ \ 5.60
            & 47.09 $\pm$16.52	& 42.01$\pm$17.92	
    & 18.47$\pm$3.11	& 17.36$\pm$2.65
\\\midrule
    \method{DGI/InfoGraph}
    & 59.42$\pm$18.82	& 56.24$\pm$24.58	
    & 68.24$\pm$15.11	& 65.35$\pm$16.16	
    & 18.30$\pm$\ \ 9.23	& 16.84$\pm$6.74
        & 34.73$\pm$16.38	& 20.67$\pm$16.65
            & 51.01$\pm$15.60	& 45.13$\pm$16.04	
    & 18.20$\pm$2.85	& 17.39$\pm$2.44
\\
    \method{GraphCL}
    & 58.53$\pm$16.64	& 55.46$\pm$21.30	
    & 66.59$\pm$15.12	& 64.67$\pm$15.25	
    & 18.40$\pm$\ \ 9.23	& 16.93$\pm$6.24
        & 33.17$\pm$13.49	& 21.75$\pm$12.03
            & 50.62$\pm$15.78	& 44.78$\pm$16.21	
    & 18.32$\pm$2.85	& 17.46$\pm$2.38
\\\midrule
    \method{CPT-HG}
    & 62.57$\pm$15.40	& 58.47$\pm$20.10	
    & 70.63$\pm$13.46	& 67.03$\pm$12.58	
    & 19.00$\pm$\ \ 8.55	& 17.46$\pm$5.51
        & 35.78$\pm$19.57	& 27.06$\pm$18.89
            & 53.67$\pm$14.98	& 46.98$\pm$15.52	
    & 18.95$\pm$2.74	& 17.94$\pm$2.13
\\
    \method{HeCo}
    & 63.32$\pm$15.14	& 59.13$\pm$17.46	
    & 70.74$\pm$11.60	& 67.63$\pm$12.35	
    & 19.42$\pm$\ \ 8.13	& 18.17$\pm$6.18
        & 35.42$\pm$15.96	& 27.37$\pm$14.45
            & 53.78$\pm$13.21	& 47.51$\pm$15.28	
    & 19.30$\pm$2.68	& 18.54$\pm$2.38
\\\midrule
    \method{GPPT}
    & 54.89$\pm$10.89	& 51.67$\pm$12.24
    & 61.47$\pm$\ \ 9.23	& 63.23$\pm$10.67	
    & 17.58$\pm$\ \ 7.67	& 16.57$\pm$5.19
    & -  & -
    & - & -
    & - & -\\
    \method{GraphPrompt}
    & 65.67$\pm$13.69	& 61.10$\pm$14.85	
    & 73.73$\pm$\ \ 9.94	& 69.26$\pm$10.63	
    & 19.86$\pm$\ \ 6.07	& 18.18$\pm$4.43
        & 36.18$\pm$15.76	& 27.70$\pm$15.41
            & 56.68$\pm$11.62	& 49.08$\pm$13.69	
    & 19.77$\pm$2.44	& 18.55$\pm$1.81
\\\midrule
    \method{\model}
    & \underline{68.23}$\pm$11.35	& \underline{63.57}$\pm$13.09	
    & \underline{77.69}$\pm$\ \ 9.70 & \underline{73.22}$\pm$10.95	
    & \underline{21.26}$\pm$\ \ 7.00 & \underline{19.81}$\pm$4.83
        & \underline{37.57}$\pm$15.08	& \underline{28.90}$\pm$14.88
            & \underline{58.49}$\pm$10.68 & \underline{47.75}$\pm$\ \ 8.80	
    & \underline{20.91}$\pm$2.82 & \underline{19.86}$\pm$2.14
\\
    \method{\pmodel}
    & \textbf{74.04}$\pm$12.49 & \textbf{74.47}$\pm$14.28	
    & \textbf{82.31}$\pm$10.67 & \textbf{77.44}$\pm$12.05	
    & \textbf{22.64}$\pm$\ \ 7.94	& \textbf{21.08}$\pm$5.72
        & \textbf{39.72}$\pm$15.08 & \textbf{32.96}$\pm$18.07
            & \textbf{60.53}$\pm$11.40 & \textbf{52.02}$\pm$13.95	
    & \textbf{21.70}$\pm$3.32	& \textbf{20.91}$\pm$3.29
\\\midrule
    \bottomrule
        \end{tabular}}
\end{table*}

\subsection{Experimental Setup}
\stitle{Datasets.}
We conduct experiments on three benchmark datasets.
(1) \emph{ACM} serves as a citation network, which comprises papers from five conferences, classified into three distinct categories: Database, Wireless Communication, and Data Mining.
(2) \emph{DBLP} serves as an all-encompassing bibliographic database housing computer science research papers and proceedings.
(3) \emph{Freebase} \cite{bollacker2008freebase} is a functional and scalable tuple database utilized for organizing comprehensive human knowledge in a structured manner. 
For all these datasets, we employ the same raw data as Simple-HGN \cite{lv2021we}.
We provide a summary of these datasets in Table~\ref{table.datasets}. 

\stitle{Baselines.}
We assess the performance of \model\ against state-of-the-art approaches from five primary categories as outlined below.

(1) \emph{End-to-end homogeneous graph neural networks}: GCN \cite{kipf2016semi} and GAT \cite{velivckovic2018graph}. These homogeneous GNNs leverage the central operation of neighborhood aggregation to iteratively collect messages from neighboring nodes, operating in an end-to-end manner.
(2) \emph{End-to-end heterogeneous graph neural networks (HGNNs)}: Simple-HGN \cite{lv2021we} and HAN \cite{wang2019heterogeneous}. Compared to homogeneous GNNs, these HGNNs incorporate heterogeneity through heterogeneous neighborhood aggregation \wrt edge types or meta-paths.
(3) \emph{Homogeneous Graph pre-training models}: DGI/InfoGraph\footnote{Original DGI only works at the node level, while InfoGraph extends it to the graph level. In our experiments, we use DGI for NC, and InfoGraph for GC.} \cite{velivckovic2018deep,Sun2020InfoGraph} and GraphCL \cite{you2020graph}. These approaches follow the ``pre-train, fine-tune'' paradigm. Specifically, they first pre-train the GNN models to exploit the intrinsic properties of the graphs, and 
subsequently fine-tune the pre-trained weights on downstream tasks to adapt to task-specific labels. 
(4) \emph{Heterogeneous Graph pre-training models}: CPT-HG \cite{jiang2021contrastive} and HeCo \cite{wang2021self}, which also follow the ``pre-train, fine-tune'' paradigm. However, they design heterogeneous tasks to learn heterogeneity information during pre-training.
(5) \emph{Graph Prompt Models}: GPPT \cite{sun2022gppt} and GraphPrompt \cite{liu2023graphprompt}. They employ LP for pre-training and unify downstream tasks with the pre-training task using a consistent template. It is worth noting that GPPT exclusively aligns the downstream NC with their pre-training task. (See Appendix B for more details).

We highlight that other few-shot methods on graphs, such as Meta-GNN \cite{zhou2019meta}, AMM-GNN \cite{wang2020graph}, RALE \cite{liu2021relative}, VNT \cite{DBLP:conf/kdd/TanGDL23} and ProG \cite{DBLP:conf/kdd/SunCLLG23}, rely on a meta-learning paradigm \cite{finn2017model}. Thus, they are not applicable to our setting since they need a large amount of labeled data in their base classes for meta-training. Conversely, our approach exclusively employs label-free graphs for pre-training.

\stitle{Settings and parameters.}
To assess if our framework generalizes well to different downstream tasks, we focus on few-shot NC and GC following previous work \cite{liu2023graphprompt}. To further test the robustness of our approach, we also adopt few-shot LP as a downstream task, although it does not have a typical few-shot setting. Hence, we will present the results on NC and GC here, and defer the results of LP to {{Appendix D}}.
Furthermore, we name our model \model\ and \pmodel: \model\ deals with the target scenario where pre-training graphs are homogeneous, while \pmodel\ copes with the scenario where pre-training graphs are heterogeneous (by also applying the graph template in pre-training); see Fig.~\ref{fig.intro-motivation}(a). 

The downstream tasks adhere to a $k$-shot learning setting. 
Further elaboration on the task construction process will be provided when presenting the results of each kind of task in Sect.~\ref{sec:expt:perf}.
We adopt MicroF and MacroF \cite{scikit-learn, lv2021we} as the evaluation metrics.

For hyperparameter settings and other implementation details about the baselines and \model, 
see {Appendix C}.  

\begin{figure}[t]
\centering
\includegraphics[width=1.0\linewidth]{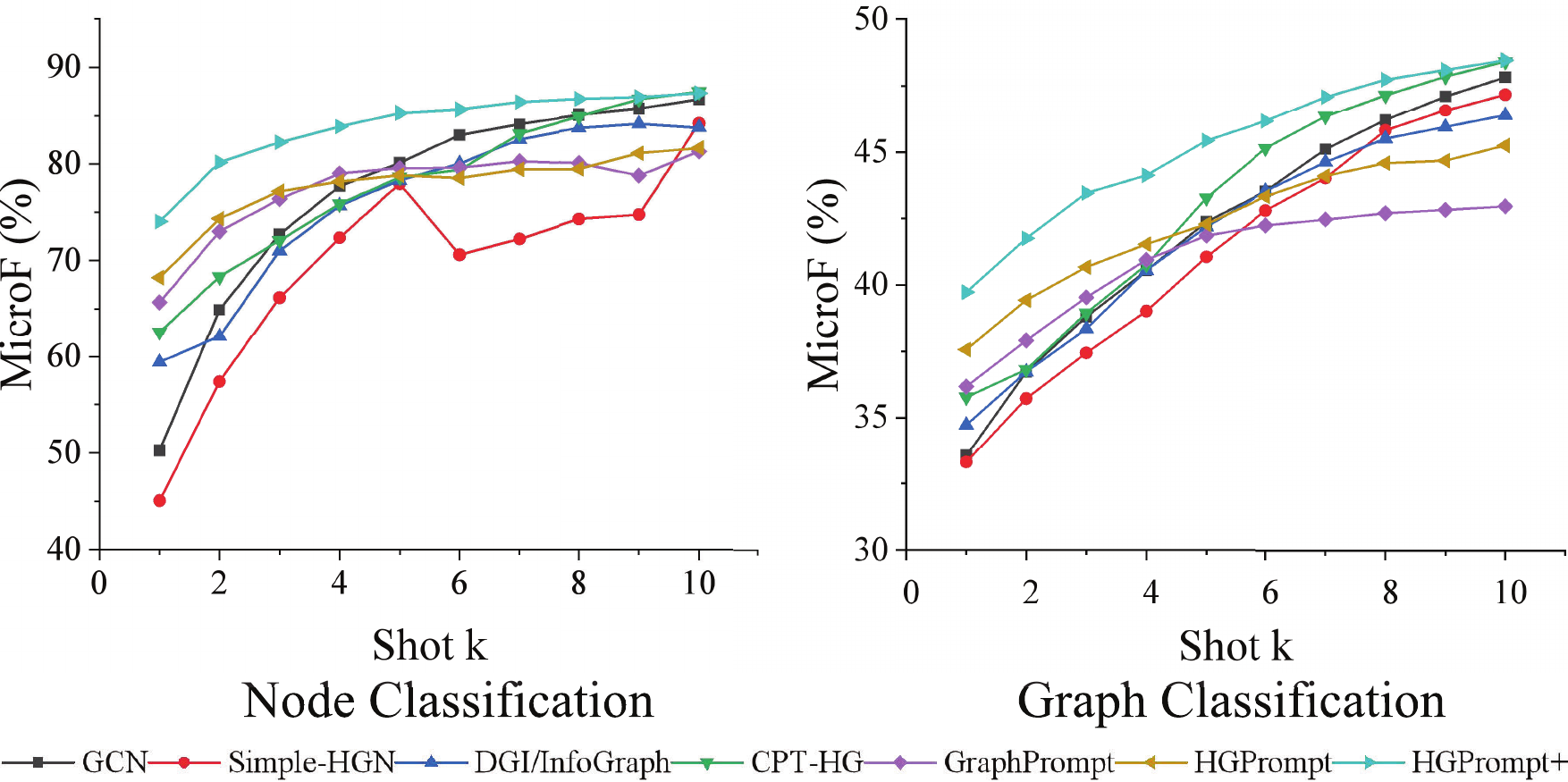}
\vspace{-4mm}
\caption{Impact of shots on NC and GC tasks on \emph{ACM}.}
\label{fig.few-shot}
\end{figure}


\begin{table*}[tb]
    \centering
    \small
    \addtolength{\tabcolsep}{1mm}
    \caption{Variants used in ablation study, and corresponding results in MicroF (\%) on node and graph classification.}
    \vspace{-1mm} 
    \label{table.ablation-illu}%
    \resizebox{0.9\linewidth}{!}{%
    \begin{tabular}{@{}l|cccc|ccc|ccc@{}}
    \toprule
    \multirow{2}*{Methods}
    & Graph template & Heterogeneity & Graph template & Feature &\multicolumn{3}{c|}{Node classification} &\multicolumn{3}{c}{Graph classification}\\
    & (pre-training) &prompt &(downstream) &prompt & ACM & DBLP & Freebase & ACM & DBLP & Freebase\\
    \midrule\midrule
    \method{Variant 1} 
    & $\times$ & $\times$ & $\times$ & $\times$ &55.92 &56.49 &17.43 &33.92 &40.47 &16.78\\ 
    \method{Variant 2}
    & $\times$ & $\times$ & $\times$ & $\checkmark$ &61.10 &69.26 &18.18 &35.18 &56.68 &19.77\\
    \method{Variant 3}
    & $\times$ & $\times$ & $\checkmark$ & $\checkmark$ &62.34 &70.77 &18.63 &35.70 &56.23 &19.03\\
    \method{\model}
    & $\times$ & $\checkmark$ & $\checkmark$ & $\checkmark$ &\textbf{63.57} &\textbf{73.22} &\textbf{19.81} &\textbf{37.57} &\textbf{58.49} &\textbf{20.91}\\\midrule
    \method{Variant 1+} 
    & $\checkmark$ & $\times$ & $\times$ & $\times$ &61.79 &60.71 &18.09 &35.48 &48.56 &17.81\\ 
    \method{Variant 2+}
    & $\checkmark$ & $\times$ & $\times$ & $\checkmark$ &68.47 &72.05 &19.88 &36.85 &58.24 &20.85\\
    \method{Variant 3+}
    & $\checkmark$ & $\times$ & $\checkmark$ & $\checkmark$ &71.37 &74.36 &20.73 &37.60 &59.08 &20.32\\
    \method{\pmodel}
    & $\checkmark$ & $\checkmark$ & $\checkmark$ & $\checkmark$ &\textbf{74.47} &\textbf{77.44} &\textbf{21.08} &\textbf{39.72} &\textbf{60.53} &\textbf{21.70} \\
    \bottomrule
    \end{tabular}}
    
\end{table*}

\subsection{Performance Evaluation}\label{sec:expt:perf}
For both few-shot NC and GC tasks, 
we first evaluate a fixed-shot setting, and subsequently vary the number of  shots to observe the performance trends. 

\stitle{Few-shot node classification.}
Following the typical $k$-shot setup \cite{wang2020graph, liu2021relative, liu2023graphprompt}, we randomly generate 100 one-shot tasks for model training and validation. 
The results of one-shot NC are presented in Table~\ref{table.node-graph-classification}. We make the following observations. 
(1) \model\ surpasses all baselines on the three datasets, indicating the efficacy of \model\ in bridging homogeneous and heterogeneous graphs across pre-training and downstream tasks. 
(2) \pmodel\ outperforms \model, demonstrating the benefit of further leveraging heterogeneity in pre-training (assuming made available) via graph templates. Similarly, CPT-HG and HeCo also make use of heterogeneity in pre-training, and perform better than their homogeneous counterparts DGI/InfoGraph and GraphCL. However, CPG-HG and HeCo still lag behind \model\ and \pmodel\ significantly despite an already unified graph format (i.e., heterogeneous graphs in both pre-training and downstream tasks), further implying the importance of the task template and our dual-prompt design in few-shot settings. 
(3) HAN outperforms other supervised methods because it leverages meta-paths, which are pre-defined using domain knowledge. Also note that GAT and Simple-HGN often perform poorly due to their large model size, which are not intended for few-shot settings.

\stitle{Few-shot graph classification.}
Following previous work \cite{lu2021learning}, we create a set of graphs by gathering ego-networks centered on the target nodes (i.e., those with class labels) in each dataset. Each ego-network is assigned the same label as its corresponding ego-node. 
Again, we randomly generate 100 one-shot GC tasks for model training and validation. 
We report the results of one-shot GC in Table~\ref{table.node-graph-classification}.
We observe similar patterns to those on the NC results, showing robustness of our approach on both node- and graph-level tasks.


\stitle{Performance with different shots.}
To further investigate the impact of shots, we vary the number of shots for both NC and GC tasks, and benchmark against several competitive baselines in Fig.~\ref{fig.few-shot} on \emph{ACM} (see {Appendix D} for other datasets).
We make the following observations.
(1) \model\ outperforms the baselines when very limited labeled data are given (\eg, fewer than 5 shots), and
remains competitive when more shots are available. Note that CPT-HG utilizes heterogeneous information in pre-training, and hence has an advantage to \model, and \pmodel\ serves as a better comparison. (2) 
\pmodel\ outperforms all baselines even when increasing to 10 shots, by also making use of the heterogeneity via graph templates in pre-training. 
(3) Generally, the performance advantages of \model\ and \pmodel\ diminish as more shots are used, which is expected as they are especially designed for label-scarce settings.

\subsection{Model Analyses}

\stitle{Ablation study.}
To study the impact of each component within our model, we undertake an ablation study. We modify \model\ against design variations associated with the key components of dual-template and dual-prompt, as shown in Table~\ref{table.ablation-illu}.
Note that for all variants, the task template is always applied as that is the basis of our framework. Furthermore, certain constraints exist, \eg, a heterogeneity prompt is only meaningful if the graph template is applied downstream.
%
%
%
%
%
%

The results in Table~\ref{table.ablation-illu} show that various components are useful, as follows.
(1) The absence of a heterogeneity prompt diminishes performance. This is evident when contrasting \model\ with Variant 3 and \pmodel\ with Variant 3+, respectively, demonstrating the need to bridge the heterogeneity differences across tasks.
(2) Omitting graph template for downstream tasks similarly impacts outcomes adversely, given the generally better performance of Variant 3 against Variant 2, and that of Variant 3+ against Variant 2+.
(3) Applying graph templates during pre-training yields clear benefits. Notably, all variants annotated with ``+" 
consistently outperform their counterparts without ``+'', demonstrating that graph templates can also leverage heterogeneous information in pre-training effectively.
(4) Variants 1/1+ typically achieve the least optimal performance, underscoring the necessity of graph templates and dual-prompt.
\begin{table}[tbp]
    \centering
    \small
     \addtolength{\tabcolsep}{-1mm}
    \caption{Evaluation of different backbones for NC and GC tasks on \emph{ACM}. \textsc{Supervised} means end-to-end training.}
    \vspace{-1mm} 
    \label{table.backbone}%
    \resizebox{1\linewidth}{!}{%
    \begin{tabular}{@{}l|l|cc|cc@{}}
    \toprule
    \multirow{2}*{Backbone} & \multirow{2}*{Method} & \multicolumn{2}{c|}{Node Classification} & \multicolumn{2}{c}{Graph Classification}\\ 
    & & MicroF (\%) & MacroF (\%) & MicroF (\%) & MacroF (\%)\\
    \midrule\midrule
    \multirow{4}{*}{GCN} 
    & \method{Supervised} & 50.26 & 44.21 & 33.56	&17.72\\
    & \method{GraphPrompt} & 65.67 & 61.10 & 36.18	&27.70\\
    & \method{\model} & 68.23 & 63.57 & 37.57	&28.90\\
    & \method{\pmodel} & \textbf{74.04}	& \textbf{74.47}	& \textbf{39.72}	&\textbf{32.96}\\\midrule
    
    \multirow{4}{*}{GAT} 
    & \method{Supervised} & 38.50 & 28.01 & 33.52	&17.25
\\
    & \method{GraphPrompt} & 38.87 & 31.97 & 34.52	&23.44
\\
    & \method{\model} & 38.98 & 33.85 & 35.75	&24.56
\\
    & \method{\pmodel} & \textbf{45.86} & \textbf{42.50} & \textbf{37.10} &\textbf{25.33}
\\\midrule

    & \method{Supervised} & 45.57 & 40.58 & 33.31	&16.83
\\
\method{Simple-}    & \method{GraphPrompt} & 49.62 & 45.25 & 34.47	&21.64
\\
\method{HGN}    & \method{\model} &51.79 & 47.17 & 35.38	&22.83\\
    & \method{\pmodel} & \textbf{54.81}	& \textbf{51.06}	& \textbf{36.95}	&\textbf{24.15}\\
    \bottomrule
    \end{tabular}}
\end{table}

\stitle{Flexibility on backbones.}
To further assess the flexibility and robustness of \model, we examine its performance on different GNN and HGNN backbones, including \textsc{GCN}, \textsc{GAT} and \textsc{Simple-HGN}.
We report the results in Table~\ref{table.backbone} on \emph{ACM} (see {Appendix D} for other datasets).
We observe that irrespective of the backbone employed, \pmodel\ consistently emerges as the top performer, followed by \model, implying the robustness of our proposed framework. 

\stitle{Parameter efficiency.} We further investigate the number of parameters that require updating during downstream tasks. See {Appendix D} for more details.

\section{Conclusions}

In this study, we delved into the  challenge of few-shot prompt learning on heterogeneous graphs. We introduced \model, aiming to seamlessly bridge the gap between homogeneous and heterogeneous graphs. Specifically, we proposed dual-template to unify downstream tasks with pre-training irrespective of their graph heterogeneity. Then, we proposed dual-prompt to narrow the gap caused by feature and heterogeneity variations across tasks. Comprehensive evaluations on three benchmark datasets further illustrate the advantages of \model, which consistently outperforms state-of-the-art baselines.

\newpage
\section*{Acknowledgments}
This research / project is supported by the Ministry of Education, Singapore, under its Academic Research Fund Tier 2 (Proposal ID: T2EP20122-0041). Any opinions, findings and conclusions or recommendations expressed in this material are those of the author(s) and do not reflect the views of the Ministry of Education, Singapore. This work is also supported in part by the National Key Research and Development Program of China under Grant 2020YFB2103803. The authors wish to thank Dr.~Deyu Bo from Beijing University of Posts and Telecommunications for his valuable comments on this work.

\bibliography{references}


\end{document}


\maketitle

\setcounter{secnumdepth}{5}
\renewcommand{\thefigure}{\Roman{figure}}
\renewcommand{\thetable}{\Roman{table}}

\renewcommand\thesubsection{\Alph{subsection}}

\begin{figure*}[thb]
\centering
\includegraphics[width=1\linewidth]{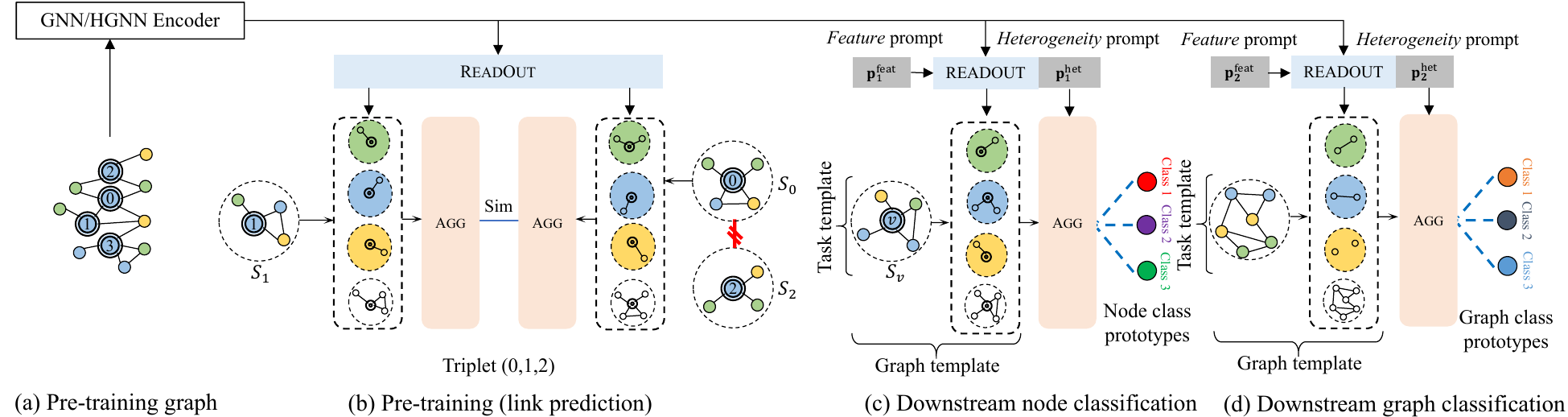}
\vspace{-4mm}
\caption{Overall framework of \model. 
(a) An input heterogeneous graph for graph pre-training.
(b) Pre-training task with link prediction on heterogeneous graphs. (c) Downstream node classification task and (d) graph classification task on heterogeneous graphs.}
\label{fig.app-framework}
\end{figure*}
\subsection{Overall Framework with Heterogeneous Graphs for Pre-training}


\model\ can be pre-trained using both homogeneous and heterogeneous graphs. While Fig.~2 in the main paper demonstrates an example of employing both graph types for pre-training, we further detail the exclusive use of heterogeneous graphs for this purpose in Fig.~\ref{fig.app-framework}. 

To elucidate, given a heterogeneous graph as input, we employ a graph template to transition it into a series of homogeneous graphs for the pre-training phase (see Figs.~\ref{fig.app-framework}(a) and (b)). We then undertake link prediction as the pre-training task (Fig.~\ref{fig.app-framework}(b)). Subsequently, both node classification and graph classification can be executed as downstream tasks on the heterogeneous graphs (Figs.~\ref{fig.app-framework}(c) and (d)).

\subsection{Further Descriptions of Baselines} \label{app.baseline}
In this section, we present more details for the baselines.
\vspace{1.5mm}

\noindent (1) \textbf{End-to-end Homogeneous Graph Neural Networks}
\begin{itemize}
\item \textbf{GCN} \cite{kipf2016semi}: GCN utilizes mean-pooling-based neighborhood aggregation to aggregate messages from neighboring nodes. 
\item \textbf{GAT} \cite{velivckovic2018graph}: GAT also relies on neighborhood aggregation for end-to-end node representation learning. It differentiates itself by assigning varied weights to neighboring nodes, adjusting their significance in the aggregation.
\end{itemize}

\noindent (2) \textbf{End-to-end Heterogeneous Graph Neural Networks}
\begin{itemize}
\item \textbf{HAN} \cite{wang2019heterogeneous}: HAN incorporates meta-paths to amalgamate different types of heterogeneity. It employs a hierarchical attention mechanism, capturing both node-level and semantic-level importance.
\item \textbf{Simple-HGN} \cite{lv2021we}: By employing GAT as the backbone, Simple-HGN further considers edge types when calculating neighbor weights during neighborhood aggregation.
\end{itemize}

\noindent (3) \textbf{Homogeneous Graph Pre-training Models}
\begin{itemize}
\item \textbf{DGI} \cite{velivckovic2018deep}: DGI is a self-supervised pre-training method on homogeneous graphs. This technique hinges on mutual information (MI) maximization, striving to maximize the estimated MI between locally augmented instances and their global representations.
\item \textbf{InfoGraph} \cite{Sun2020InfoGraph}: As an extension of DGI, InfoGraph focuses on the graph level tasks. It seeks to maximize the consistency between node and graph embeddings.
\item \textbf{GraphCL} \cite{you2020graph}: GraphCL applies diverse graph augmentations for self-supervised learning in order to exploit the structural patterns within graphs.
The objective is to maximize the agreement between different augmentations during graph pre-training.
\end{itemize}

\noindent (4) \textbf{Heterogeneous Graph Pre-training Models}
\begin{itemize}
\item \textbf{CPT-HG} \cite{jiang2021contrastive}: CPT-HG also employs contrastive learning during its pre-training phase. Positive and negative samples are distinguished based on the presence of specific type links with a target node. The goal is to enhance the similarity between the target node and positive samples while diminishing the similarity with negative samples.
\item \textbf{HeCo} \cite{wang2021self}: Adopting a cross-view contrastive mechanism, HeCo learns HIN representations from both network schema and meta-path. It uses a view mask mechanism to derive positive and negative samples.
\end{itemize}

\noindent (5) \textbf{Graph Prompt Models}
\begin{itemize}
\item \textbf{GPPT} \cite{sun2022gppt}: GPPT adopts a GNN model pre-trained through a link prediction task. It employs a prompt module to structure the downstream node classification task, aligning it with the link prediction format.
\item \textbf{GraphPrompt} \cite{liu2023graphprompt}: GraphPrompt uses subgraph similarity calculation to unify pre-training and downstream tasks such as node and graph classification. A learnable prompt is then refined during the downstream task to assimilate task-specific details.
\end{itemize}

\subsection{Implementation Details of Approaches} \label{app.parameters}

\stitle{Details of baselines.}
For all the baseline models, we use the codes officially released by their respective authors. We tune each model in line with the settings recommended in their literature to ensure optimal performance.

For baseline GCN \cite{kipf2016semi}, we employ a 3-layer architecture, and set the hidden dimensions as 64.
For GAT \cite{velivckovic2018graph}, we employ a 2-layer architecture and set the hidden dimension as 64. Besides, we apply 4 attention heads in the first GAT layer.

For HAN \cite{wang2019heterogeneous}, we employ a 2-layer GAT with 4 attention heads in the first layer, and set the hidden dimension as 64. In this paper, we use the same meta-paths as utilized in their original paper.
For Simple-HGN \cite{lv2021we}, we employ a 2-layer GAT with 4 attention heads in the first layer, and set the hidden dimension as 64. Besides, we utilize elu as the activation function.

For DGI \cite{velivckovic2018deep}, we use a 3-layer GCN as the base model, and set the hidden dimension as 64. Besides, we utilize prelu as the activation function.
For InfoGraph \cite{Sun2020InfoGraph}, we use a 3-layer GCN as the base model, and set its hidden dimensions as 64. 
For GraphCL \cite{you2020graph}, we also employ 3-layer GCN as its base model, and set the hidden dimensions as 64. In particular, we choose the augmentations of node dropping and subgraph, with default augmentation ratio 0.2.

For CPT-HG \cite{jiang2021contrastive}, we employ a 2-layer HGT with 1 attention head as its base model, and set the dimension of node representations to 64. In particular, the maximum number of negative samples from inconsistent relations and unrelated nodes are set to 100 and 200, respectively. The maximum number of queued negative samples at subgraph-level is set to 100.
For HeCo \cite{wang2021self}, we use one-layer GCN for every meta-path, and we only consider interactions between nodes of the target type and their one-hop neighbors of other types, and set the hidden dimensions as 64.

For GPPT \cite{sun2022gppt}, we utilize a 2-layer GraphSAGE as its base model, and set the hidden dimension as 64. For this based GraphSAGE, we also utilize the mean aggregator.
For GraphPrompt \cite{liu2023graphprompt}, we employ a 3-layer GCN as the base model for \emph{ACM} and \emph{Freebase}, and a 2-layer GAT for \emph{DBLP}. We set the hidden dimensions as 64. In addition, we sample 1-hop subgraph for \emph{ACM} and \emph{Freebase}, and 2-hop subgraph for \emph{DBLP}.

\stitle{Details of \model.}
For our proposed \model, we employ a 3-layer GCN as the base model for \emph{ACM} and \emph{Freebase}, and a 2-layer GAT serves as the base model for \emph{DBLP}. We set the hidden dimensions as 64. In particular, the maximum number of negative samples for pre-training is set to 1.

\subsection{Further Experiments and Analysis}\label{app.exp}

\stitle{Few-shot link prediction.}
Beyond the primary experiments detailed in our main paper, which encompass few-shot node classification and graph classification, we leverage the shared pre-trained model to execute few-shot link prediction on heterogeneous graphs as a subsequent task.

Specifically, in the context of downstream few-shot link prediction, we are provided with $k$ exemplary ``heterogeneous links'' to guide the prompt learning, namely, $k$-shot. Each of these heterogeneous links, or edges, denoted as $(a,b)$, comes with associated heterogeneous types for both nodes $a$ and $b$, as well as the edge itself. Note that while numerous other links may exist within the graph in the downstream link prediction task, their heterogeneity is not specified. Hence, they cannot be employed as supervision in the downstream prompt learning task, given the absence of heterogeneity.

For each target node, we select one of its neighbors to be the positive sample and choose 10 nodes with no connections as the negative samples. The evaluation criteria dictate that the positive sample should rank higher than the negative samples. To assess this, we employ the AUC \cite{lu2011link, zhang2018link} and NDCG \cite{han2020metapath, fu2022revisiting} metrics.
From each dataset, we randomly pick 10,000 edges dedicated to the downstream task, ensuring these edges remain concealed during pre-training. Specifically, 500 of these edges are allocated for training, another 500 for validation, with the rest for testing.
For the evaluative process, we create a $1$-shot link prediction tuple (with one positive link and ten negative links) by randomly drawing one sample from the training dataset. We also establish a validation set, maintaining the $1$-shot configuration. This sampling approach is repeatedly applied, culminating in a collection of 100 distinct few-shot tasks.

The outcomes of few-shot LP are presented in Table~\ref{table.link-prediction}, from which we have several observations.
Initially, it is evident that our proposed \model\ consistently outclasses all baselines on \emph{ACM}, \emph{DBLP}, and \emph{Freebase}. This further underscores \model's prowess in bridging the gap between homogeneous and heterogeneous graphs.
Subsequently, even though \pmodel\ still outperforms \model, the advantage is not as prominent as it is in NC and GC. This is primarily because there is not a discernible discrepancy between the objectives of pre-training and the downstream task since they are both LP. The sole distinction revolves around the presence or absence of heterogeneity which is adeptly addressed by \model.

\begin{table}[tbp] 
    \centering
    \small
     \addtolength{\tabcolsep}{-1mm}
    \caption{Evaluation on link prediction.}
    \vspace{-3mm} 
    \label{table.link-prediction}%
    \resizebox{1\linewidth}{!}{%
    \begin{tabular}{@{}l|cc|cc|cc@{}}
    \toprule
   \multirow{2}*{Methods} &\multicolumn{2}{c|}{ACM}
   & \multicolumn{2}{c|}{DBLP} & \multicolumn{2}{c}{Freebase}\\ 
      & AUC (\%)& NDCG (\%)& AUC (\%)& NDCG (\%) & AUC (\%)& NDCG (\%)
      \\\midrule\midrule
    \method{GCN} 
        & 22.30$\pm$13.94 & 31.56$\pm$3.95
    & 49.37$\pm$11.76 & 43.40$\pm$5.95 
    & 75.58$\pm$5.42 & 65.97$\pm$4.13\\ 
    \method{GAT} 
    & 21.72$\pm$8.18 & 31.14$\pm$2.65
    & 37.08$\pm$12.15 & 36.89$\pm$5.09 
    & 81.54$\pm$3.06 & 70.63$\pm$2.42\\\midrule
    \method{Simple-HGN}
        & 22.70$\pm$11.57	& 31.81$\pm$3.58
    & 43.60$\pm$10.24	& 39.66$\pm$4.72	
    & 71.50$\pm$3.76	& 61.37$\pm$2.83\\
    \method{HAN}
        & 35.29$\pm$9.97	& 35.39$\pm$2.98
    & 56.88$\pm$12.35	& 49.88$\pm$5.89	
    & 83.85$\pm$6.02 & 71.76$\pm$4.86\\\midrule
    \method{DGI}
        & 46.67$\pm$15.74	& 40.81$\pm$5.72
    & 64.88$\pm$6.93	& 54.00$\pm$2.82	
    & 80.42$\pm$5.03	& 69.71$\pm$3.38\\
    \method{GraphCL}
        & 48.78$\pm$13.58	& 41.49$\pm$5.20
    & 66.51$\pm$8.09	& 55.54$\pm$3.53	
    & 81.46$\pm$4.82 & 74.31$\pm$5.19\\\midrule
    \method{CPT-HG}
        & 52.33$\pm$11.98	& 43.34$\pm$3.79
    & 67.31$\pm$6.15	& 56.03$\pm$4.94	
    & 85.81$\pm$7.30	& 76.29$\pm$5.86\\
    \method{HeCo}
        & 52.14$\pm$9.31	& 43.29$\pm$3.20
    & 67.35$\pm$5.73	& 56.60$\pm$4.40	
    & 86.35$\pm$5.17	& 76.65$\pm$4.21\\\midrule
    \method{GraphPrompt}
        & 54.68$\pm$6.39	& 45.13$\pm$3.14
    & 70.41$\pm$2.85	& 58.79$\pm$1.60	
    & 87.21$\pm$3.73	& 79.55$\pm$4.02\\\midrule
    \method{\model}
        & \underline{57.17}$\pm$5.18	& \underline{46.14}$\pm$2.78
    & \underline{72.81}$\pm$2.19 & \underline{60.21}$\pm$1.32	
    & \underline{88.96}$\pm$2.43 & \underline{80.46}$\pm$2.91\\
    \method{\pmodel}
        & \textbf{58.08}$\pm$5.91 & \textbf{46.88}$\pm$3.27
    & \textbf{73.64}$\pm$2.94 & \textbf{60.80}$\pm$1.33	
    & \textbf{90.04}$\pm$3.90	& \textbf{81.43}$\pm$3.74\\\midrule
    \bottomrule
        \end{tabular}}
    \vspace{-1mm}
\end{table}

\begin{figure}[t]
\centering
\includegraphics[width=1\linewidth]{figures/few-shot - dblp.pdf}
\vspace{-2mm}
\caption{Impact of shots on DBLP.}
\label{fig.few-shot-dblp}
\end{figure}

\stitle{Performance with different shots.}
We also tune the number of shots for few-shot node and graph classification tasks to show its impact. Specifically, we experiment with a range of shots, spanning from 1 to 10, and benchmarked against some competitive baselines, including GCN, Simple-HGN, GraphCL, CPT-HG, and GraphPrompt.
The task setting remains consistent with our previously established settings for both few-shot NC and GC. For our empirical studies, we focus on the \emph{ACM} and \emph{DBLP}. Notably, \emph{Freebase} was excluded since it lacks a sufficient number of samples for certain classes beyond 3-shot learning. The result on \emph{ACM} and analysis are shown in the main paper (Sect.~5.2). The result on \emph{DBLP} is illustrated in Fig.~\ref{fig.few-shot-dblp}.
Consistent with the results on \emph{ACM}, \model\ excels over baselines when only a limited amount of labeled data is available. Meanwhile, \pmodel\ consistently outperforms all baselines across multiple-shot scenarios.

\stitle{Performance based on different backbones.}
The evaluation is conducted on \emph{DBLP} and \emph{Freebase} datasets for NC and GC, as illustrated in Table~\ref{table.backbone}. Our observations from this study align with those on \emph{ACM} as illustrated in Sect.~5.3 in the main paper.

\begin{table}[tbp]
    \centering
    \small
    \addtolength{\tabcolsep}{-1mm}
    \caption{\model\ and \pmodel's performance with various backbones. MiF is short for MicroF (\%) and MaF is short for MacroF (\%).}
    \vspace{-3mm} 
    \label{table.backbone}%
    \resizebox{\linewidth}{!}{%
    \begin{tabular}{@{}l|l|cccc|cccc@{}}
    \toprule
    \multirow{3}*{Backbones} & \multirow{3}*{Methods} &\multicolumn{4}{c|}{Node classification} &\multicolumn{4}{c}{Graph classification}\\
    & &\multicolumn{2}{c}{DBLP} & \multicolumn{2}{c|}{Freebase} & \multicolumn{2}{c}{DBLP} & \multicolumn{2}{c}{Freebase}\\ 
    & & MiF & MaF & MiF & MaF & MiF & MaF & MiF & MaF\\
    \midrule\midrule
    \multirow{4}{*}{GCN} 
    & \method{Supervised} & 51.36 & 48.62 & 17.11 & 15.35  &38.12	&35.01	&17.38	&16.50\\
    & \method{GraphPrompt} & 65.79 & 63.32 & 19.86 & 18.18 &43.58	&40.23	&19.77	&18.55\\
    & \method{\model} & 75.72 & 71.08 & 21.26 & 19.81 &44.23	&41.68	&20.91	&19.86\\
    & \method{\pmodel} & 79.98	& 76.84	& \textbf{22.64} & \textbf{21.08} &53.29	&46.52	&\textbf{21.70}	&\textbf{20.91}\\
    \midrule
    
    \multirow{4}{*}{GAT} 
    & \method{Supervised} & 65.04	& 62.83	& 17.93	& 16.51 &46.47	&40.33	&16.30	&16.01\\
    & \method{GraphPrompt}  & 73.73 &69.26 & 18.39 & 15.54 &52.83	&45.68	&17.54	&17.42\\
    & \method{\model} & 77.69 & 73.22 & 20.22 & 18.26 &54.49	&46.75	&18.86	&18.52\\
    & \method{\pmodel} & \textbf{82.31} & \textbf{77.44} & 21.48 & 20.21 &\textbf{60.53} &\textbf{52.02} &19.92	&19.34\\
    \midrule

    \multirow{4}{*}{SHGN} 
    & \method{Supervised}  & 59.74 & 57.34 & 17.85 & 16.00 &44.56	&39.52	&17.83	&16.71 \\
    & \method{GraphPrompt} & 61.93 & 58.19 & 19.06 & 17.80  &49.23	&43.84	&19.58	&18.69 \\
    & \method{\model} & 62.81 & 60.95 & 19.73 & 18.26  &50.87	&45.02	&20.33	&19.34\\
    & \method{\pmodel} & 66.05	& 64.08	&21.50 & 19.74 &57.02 &49.97	&20.58	&20.45 \\
    \midrule
    \end{tabular}}
\end{table}

\stitle{Parameter efficiency.}
We further analyze the number of parameters necessitated for optimization during the downstream task across various representative models. To elucidate, we spotlight the node classification task as a representative example, though findings from other tasks largely mirror these conclusions. These statistics are presented in Table~\ref{table.parameters-num}.
Notably, as both GCN and Simple-HGN adopt an end-to-end approach, they invariably require updates to a substantial portion of parameters---essentially, all parameters within their respective models, rendering them the most intensive.
On the other hand, models like DGI and CPT-HG, which embrace a ``pre-train, fine-tune'' paradigm, only mandate updates to the downstream classifier. This results in a significant reduction in the number of parameters subject to modification.
It is noteworthy that our proposed \model\ not only eclipses other baselines in performance but also necessitates updates to a considerably leaner set of parameters, second only to GraphPrompt. This efficiency is especially commendable considering the remarkable performance boost achieved with such a modest parameter increment.

\begin{table}[!t]
    \centering
    \renewcommand{\arraystretch}{1.2}
    \addtolength{\tabcolsep}{-3pt}
    \caption{Comparison for parameters on downstream node classification.}
    \resizebox{0.6\linewidth}{!}{%
    \begin{tabular}{@{}l|c|c|c@{}}
    \toprule
    \multirow{2}*{Methods} & \multicolumn{3}{c}{Datasets} \\
    & ACM & DBLP & Freebase \\ \midrule\midrule
    GCN & 70,496 & 1,676,932 & 11,531,463 \\
    Simple-HGN & 1,264,838 & 2,070,600 & 12,177,806 \\
    DGI & 192 & 256 & 448 \\
    CPT-HG & 192 & 256 & 448 \\
    GraphPrompt & 64 & 64 & 64 \\ 
    \model & 67 & 68 & 71 \\ \bottomrule
    \end{tabular}}
    \label{table.parameters-num}
\end{table}


\bibliography{references}